\documentclass[11pt,a4paper]{article}
\usepackage[hyperref]{eacl2021}
\usepackage{times}
\usepackage{latexsym}
\usepackage{graphicx}
\usepackage{listings}
\usepackage{enumitem}
\usepackage{clrscode}
\usepackage{amsmath,amssymb}
\usepackage{bbm}
\usepackage{hyperref}
\usepackage{booktabs}
\usepackage{multirow}
\usepackage{listings}
\usepackage{subcaption}

\usepackage[hang,flushmargin]{footmisc}

\usepackage{cleveref}
\crefname{table}{Table}{Tables}
\crefname{figure}{Figure}{Figures}
\crefname{appendix}{Appendix}{Appendices}
\crefname{section}{Section}{Sections}

\usepackage{microtype}
\aclfinalcopy 
\def\aclpaperid{1116} 

\makeatletter
\newcommand*\iftodonotes{\if@todonotes@disabled\expandafter\@secondoftwo\else\expandafter\@firstoftwo\fi}  
\makeatother

\title{Gender and Racial Fairness\\
in Depression Research using Social Media}

\author{Carlos Aguirre, Keith Harrigian, Mark Dredze  \\
  Johns Hopkins University \\
  \texttt{caguirr4@jhu.edu, kharrigian@jhu.edu, mdredze@cs.jhu.edu} \\
  }

\date{}

\begin{document}
\maketitle 
\begin{abstract}
Multiple studies have demonstrated that behavior on internet-based social media platforms can be indicative of an individual's mental health status. The widespread availability of such data has spurred interest in mental health research from a computational lens. While previous research has raised concerns about possible biases in models produced from this data, no study has quantified how these biases actually manifest themselves with respect to different demographic groups, such as gender and racial/ethnic groups. Here, we analyze the fairness of depression classifiers trained on Twitter data with respect to gender and racial demographic groups. We find that model performance systematically differs for underrepresented groups and that these discrepancies cannot be fully explained by trivial data representation issues. Our study concludes with recommendations on how to avoid these biases in future research.
\end{abstract}


\section{Introduction}
Work from \citet{dechoudhury2013predicting} and \citet{coppersmith-etal-2014-quantifying}, showing that an individual's mental health can be evaluated based on the language they generate on social media platforms, has served as the basis for a substantial amount of computational research over the last decade. Subsequent studies have examined an even wider range of mental health conditions, social media platforms, and types of online behavior at both the individual and population level \citep{coppersmith-etal-2015-clpsych, lynn-etal-2018-clpsych, De-Choudhury-etal-2016-discovering}. Vast potential for societal benefits underlies this work, as conservative estimates suggest that 8.1\% of American adults suffer from major depressive disorder at any given time and up to 16.2\% of individuals will experience at least one major depressive episode during their lifetime \citep{kessler2003epidemiology,CDC,Hasin-2018-epidemiology}. Mental health services are transitioning to online mediums at a rapid pace, with the recent COVID-19 pandemic dramatically further accelerating this trend \cite{zhou2020role, ohannessian2020global}. 
Thus, analysis of online language may play a key role in mental health treatment in the future.

Nonetheless, care must be taken to understand potential biases inherent in this research before any technologies are deployed in a clinical setting. For instance, previous work has found that Black and Hispanic/Latinx individuals are less likely to be treated for depression than White individuals \citep{simpson2007racial}. Possibly a result of this underlying bias, recent studies of the US population have concluded that baseline rates of depression vary depending on demographics \citep{CDC, Hasin-2018-epidemiology} --- major depressive disorder was found to be more prevalent in females and White adults. Yet, it remains unclear whether these supposed differences in depression prevalence between gender and racial/ethnic demographic groups are the result of measurement error or other confounders. Various psychological studies have found mental health disorders, including depression, may manifest differently depending on cultural background and thus make uniform diagnosis a difficult proposition \citep{blanchard2020reexamining, henrich_heine_norenzayan_2010}.
These ambiguities were highlighted by recent computational research from \citet{amir-etal-2019-mental}, which found predictive rates of depression inferred using classifiers for social media data to not match previous US depression estimates. Indeed, the authors actually find that Black and Hispanic/Latinx individuals are \emph{more} likely to be affected by depression than White individuals.

Additionally, NLP and other data-driven algorithms have been shown to suffer from {\it content} biases; that is, undesirable group-wise differences with respect to protected groups, such as race/ethnicity or gender \citep{johannsen2015cross, hovy2015tagging, bolukbasi2016man,gonen2019lipstick,rudinger2018gender}.
Therefore, in consideration of the social impact of NLP research \cite{hovy2016social}, in the area of mental health content analysis it stands to reason that we should also look for {\it population} biases, as they pertain to protected groups, and the way these might affect NLP algorithms' fairness.

Previous research has utilized user demographics within social media mental health studies to construct control groups \citep{coppersmith-etal-2014-quantifying}, to enhance classifier performance through additional features \citep{pietro-etal-2015-role}, and to analyze trends amongst specific populations \citep{De-Choudhury-2014-Characterizing}.
In an attempt to preemptively address {\it population} biases, \citet{amir-etal-2019-mental} proposed a cohort-based sampling approach to collect representative measures of wellness amongst the general population. 
However, as noted in a recent literature reviews \citep{chancellor2020methods, harrigian2020state}, no previous computational mental-health study has accounted for differences in population-level depression rates nor explored performance variations across demographic subgroups at training time. Therefore, little is known about the fairness of these automated systems. Are models trained for mental health fair across demographic groups? Are current datasets demographically representative? If bias exists, what is its source?

In this study, we analyze two common depression-inference datasets and explore the susceptibility of different computational methods to demographic biases.
We find that existing datasets are {\bf not demographically representative} and that, without accounting for this, we find degradation in model performance for underrepresented groups. We explore the possible sources of this bias and conclude with recommendations for future research that may address these issues.

\section{Mental Health and Social Media}

Challenges obtaining mental health annotations for social media data have thus far constrained the size and quality of existing datasets. For instance, manual annotation of mental health status generally requires expert domain knowledge, while the sensitive nature of such annotations limit multi-institutional data sharing \citep{arseniev2018type}. 
Consequently, most datasets rely on labels based on behavioral proxies or self-reported diagnoses, which more easily scale, but introduce problematic \textit{self-disclosure} bias and label noise.
Furthermore, as our understanding of mental health is continually evolving, studies have used different and sometimes conflicting guidelines for annotation \cite{CDC, Hasin-2018-epidemiology}. With these challenges at the forefront of dataset curation, issues surrounding demographic balance and representation have been largely kicked down the road of the research domain.

Challenges accounting for demographics go beyond the computational research space and are well-illustrated by disparities between two recent surveys of depression prevalence. The Centers for Disease Control and Prevention (CDC) found depression prevalence between race/ethnicity groups did not differ \citep{CDC}, while a study using the results of the National Epidemiologic Survey on Alcohol and Related Conditions III (NESARC-III) found depression to be more prevalent in White Americans versus minorities \citep{Hasin-2018-epidemiology}.

\section{Ethical Considerations}
\label{sec:ethics}

The sensitive nature of mental health research and individual demographics requires us to consider possible benefits of this study alongside its potential harms. Specifically, we must evaluate the cost-benefit trade-off of inferring and/or securing three highly-personal individual attributes: (\textit{depression diagnoses}: \citealp{benton-etal-2017-ethical}; \textit{gender identity}: \citealp{larson-2017-gender}; \textit{race/ethnicity identity}: \citealp{wooddoughty2020using}).

The potential immediate benefit of this study is a better understanding of demographic bias in computational mental health research. A potential secondary benefit is the mitigation of extant clinical treatment disparities \citep{simpson2007racial}. As mental health treatment increasingly adopts an online delivery mechanism, this research is uniquely situated to inform the development of new AI systems and public policy in the area.

However, we are cognizant of the potential harms from our work. Mental health status and demographic identities are both sensitive personal attributes that could be used to maliciously target individuals on publicly-facing online platforms. Therefore, we follow the guidelines of \citet{benton-etal-2017-ethical} and \citet{ayers2018don} on data use, storage, and distribution. All analysis was conducted on de-identified versions of data, with any identifiable information being used only during intermediate data-processing subroutines that were hidden from researcher interaction and approved by the original dataset distributors. Our study was exempted from review by our Institutional Review Board under 45 CFR § 46.104. 

To facilitate any form of statistical analysis, we also need to formalize gender and race/ethnicity. We seek a balance between the limitations of demographic inference systems\footnote{We infer gender and race/ethnicity labels using a content classifier and explore additional limitations in \cref{sec:demographer}} and alignment to demographic categories conventions used by the mental health literature \citep{CDC, Hasin-2018-epidemiology}, versus propagating demographic definitions that exacerbate existing biases towards gender and racial/ethnic minorities.
We consider the `folk conception' of gender as described in \citet{larson-2017-gender} and prominently leveraged in traditional depression research in the United States --- we use the sex categories \textit{male} and \textit{female} to denote the corresponding gender categories \textit{masculine} and \textit{feminine}.
However, many individuals do not fit in these gender categories, some present a gender online inconsistent to their true identity \citep{nilizadeh2016twitter}, and they often experience depression and other mental health conditions at a higher rate \citep{mcdonald2018social}.
For race/ethnicity labels, we consider the mutually-exclusive labeling conventions invoked by \citet{CDC} and \citet{wooddoughty2020using}: non-Hispanic White, non-Hispanic Black, non-Hispanic Asian and Hispanic/Latinx, as they are representative of the majority of racial and ethnic identities in the US.
Our racial/ethnic categories do not capture multiracial individuals or those with a race/ethnicity outside this group. 

We acknowledge these important limitations, but at the same time, there is an urgency to the questions we pose. Computational methods for monitoring mental health have already been deployed by digital surveillance companies \citep{bark}, while analytics dashboards based on these methods are gradually making their way into patients' \citep{yoo2019designing} and providers' hands \citep{yoo2020designing}. The question is: should we avoid asking these questions about current datasets because we cannot produce clear answers, or should we conduct analyses with acknowledged limitations to learn what we can about research that is already being moved into products? We firmly believe the latter. Our hope is that this paper causes researchers to carefully consider these issues, elevate the need for further work, and produce studies that go beyond our study's limitations with new data and methods. This should not be the last study on this topic; rather, we hope it is the first step which can inform further critical analyses of work in this area.

\section{Datasets}
\label{sec:datasets}
We select the task of depression inference for this study, as it is the most widely studied mental health condition in social media research \citep{harrigian2020state}. We consider two Twitter datasets: \proc{CLPsych} \cite{coppersmith-etal-2015-clpsych} and \proc{Multitask} \citep{Adrian2017MultiTask}. 

\subsection{\proc{CLPsych}}
\label{sec:clpsych}
\proc{CLPsych} was introduced by \citet{coppersmith-etal-2014-quantifying} and subsequently used in the CLPsych 2015 shared task \citep{coppersmith-etal-2015-clpsych}. 
Tweets were publicly posted between 2008 and 2013. 
Users who self-disclosed a depression diagnosis were identified using regular expressions (e.g. ``I have been diagnosed with \texttt{disorder}'') and then manually reviewed by a team of clinical and computational researchers to verify authenticity of matched disclosures. 
The control group was sampled from a random pool Twitter users so that the joint distribution of inferred age and gender attributes closely resembled that of users with self-disclosed diagnoses. 
The 3000 most recent tweets from each user (as of the original dataset collection date) were retrieved. 
To reduce ambiguity in model performance that arises due to data insufficiencies, we isolate individuals with at least $100$ tweets, leading to a final dataset size of 475 depressed individuals and their matched controls (i.e. 950 total users).

\subsection{\proc{Multitask}}
\label{sec:multitask}
\citet{Adrian2017MultiTask} constructed a Twitter dataset (\proc{Multitask}) combining a subset of \proc{CLPsych} with datasets annotated using the same procedure from \citet{coppersmith-etal-2015-adhd, glen-leary}. In addition to an expanded number of unique individuals (1400 depression, 1400 control), \proc{Multitask} also boasts a more robust historical timeline of tweets for each user.

\section{Demographic Labels}
\label{sec:demographer}
Only age and gender \citep{schwartz2013personality} attributes are available in the originally distributed form of \proc{CLPsych} and \proc{Multitask}, both of which were inferred using now-outdated models. All identifying metadata was either redacted or obfuscated to preserve the privacy of individuals in these datasets. Accordingly, we are  confronted immediately by the challenge of securing accurate demographic information to facilitate a robust analysis of any potential gender and racial/ethnic biases. Fortunately, this problem has been tackled using a multitude of different techniques across multiple studies specific to mental health \citep{yazdavar2020multimodal, amir2017quantifying, pietro-etal-2015-role, coppersmith-etal-2015-adhd} and social media applications in general  \citep{volkova-etal-2014-inferring, burger-etal-2011-discriminating, fink2012inferring, rao2011hierarchical}.

We obtain race labels using a unigram model from \citet{wooddoughty2020using}, who combine multiple crowd-sourced and self-reported datasets to train classifiers for 4 demographic groups in line with the CDC's conventions \cite{CDC}: non-Hispanic Asian American (A), non-Hispanic African American (B), non-Hispanic White (W) and  Hispanic/Latinx (H/L). Their classifier achieves an accuracy of 82.3\% within intrinsic evaluations and shows even more promise as high-confidence thresholds are applied. 
To validate and further reduce noise in previously-inferred gender attributes, we train a new gender inference model on data from \citet{burger-etal-2011-discriminating} using the same architecture of \citet{wooddoughty2020using}. Our classifier obtains an accuracy of 83.3\% amongst within-distribution data and outputs a distribution of inferred gender attributes that strongly aligns with that of the original datasets. 

Although each of these procedures has strong internal validity, we recognize that inference errors incurred during this stage may confound and complicate downstream analysis of demographic bias. To mitigate this potential noise, we also de-anonymize a subset of \proc{CLPsych} with the permission of \citet{coppersmith-etal-2014-quantifying} and apply name-based demographic classifiers \citep{wood-doughty-etal-2018-predicting,wooddoughty2020using} to each user's profile to obtain ``alternative'' age and race attributes.

Between our content and name based classifiers, we are afforded the opportunity to perform downstream analysis of demographic bias based on the attributes derived using the following mechanisms:
\begin{itemize}
    \itemsep0em 
    \item \textbf{High Confidence Filter:} Only considers users whose most probable demographic class based on unigram classifier has a confidence $> .95$.
    \item \textbf{Random Sampling:} Considers all available users; randomly split each individual's tweets into two independent pools so that demographic and mental health inferences are based on separate sets of data.
    \item \textbf{Name Labels:} Only considers users from \proc{CLPsych} who could be de-anonymized; demographics annotated using name-based gender \cite{wood-doughty-etal-2018-predicting} and ethnicity classifiers \citep{wooddoughty2020using}.
\end{itemize}

While we find some variation in the individual-level demographic labels when using the three techniques, the downstream mental health models perform similarly: see details in \cref{apx:HighConfidenceFilter}. For the experiments discussed below, we report results from the most computationally-efficient approach, high confidence filtering. 

\section{Analysis}
\label{sec:experimentation}
We conduct an analysis of these datasets and depression models trained on these datasets to answer the following questions:

\begin{enumerate}
    \itemsep0em 
    \item Are depression datasets demographically representative?
    \item Do depression classifiers perform similarly across demographic groups?
    \item Can we mitigate demographic biases by changing characteristics of the dataset?
    \item Do differences in features between demographic groups account for classifier biases?
\end{enumerate}

\subsection{Are depression datasets demographically representative?}
\label{sec:DemographicBalance}
Before we can empirically measure if these datasets are demographically representative, we must first establish the expected distribution of a representative dataset. 
While the demographic groups distribution should match the true population (Twitter users with depression), there are no estimates of depression prevalence on Twitter.
Thus, we use the Twitter US population as our baseline, and combine it with estimated prevalence of depression among US demographic groups.\footnote{Our study uses mental health statistics from the United States since they are extensive and widely available. However, due to data anonymization we could not filter our data based on residence in the US. Since these datasets are filtered to focus on English accounts, US accounts likely dominate.}

{\bf Methods.} \citet{CDC} found in a study of adults ($>$20yrs.) that women are almost twice as likely to be diagnosed with depression compared to men (1.89$\times$) using Patient Health Questionnaires (PHQ-9). We refer to this study as `CDC.' 
Similarly, \citet{Hasin-2018-epidemiology} used a national survey of adults ($>$18yrs.) and the DSM-5 standard for major depressive disorder (MDD) to estimate that women are almost twice as likely to be diagnosed with depression compared to men (1.86$\times$). 
We refer to this study as `NESARC.'

While there were only small incongruencies between these studies in estimated prevalence of depression as a function of gender, there were significant discrepancies between studies with respect to estimated prevalence as a function of race/ethnicity. Specifically, CDC found that rates of depression were not statistically different between groups, whereas  NESARC found a greater prevalence of depression among Whites compared to African Americans, Hispanics/Latinx and Asian Americans (1.25x).

We project these estimates of depression prevalence to the general Twitter population, where both males and females are estimated to participate equally (approximately matching the US population). While there is a slight under-representation of White individuals in Twitter compared to US population (60\% vs 64\%), Black and Hispanic/Latinx individuals are well represented \citep{wojcik_hughes_2019}.
Thus, barring slight variations, Twitter roughly mimics the demographic composition of the United States demographics fairly with respect to gender and race/ethnicity.

\begin{figure}
  \includegraphics[width=\columnwidth]{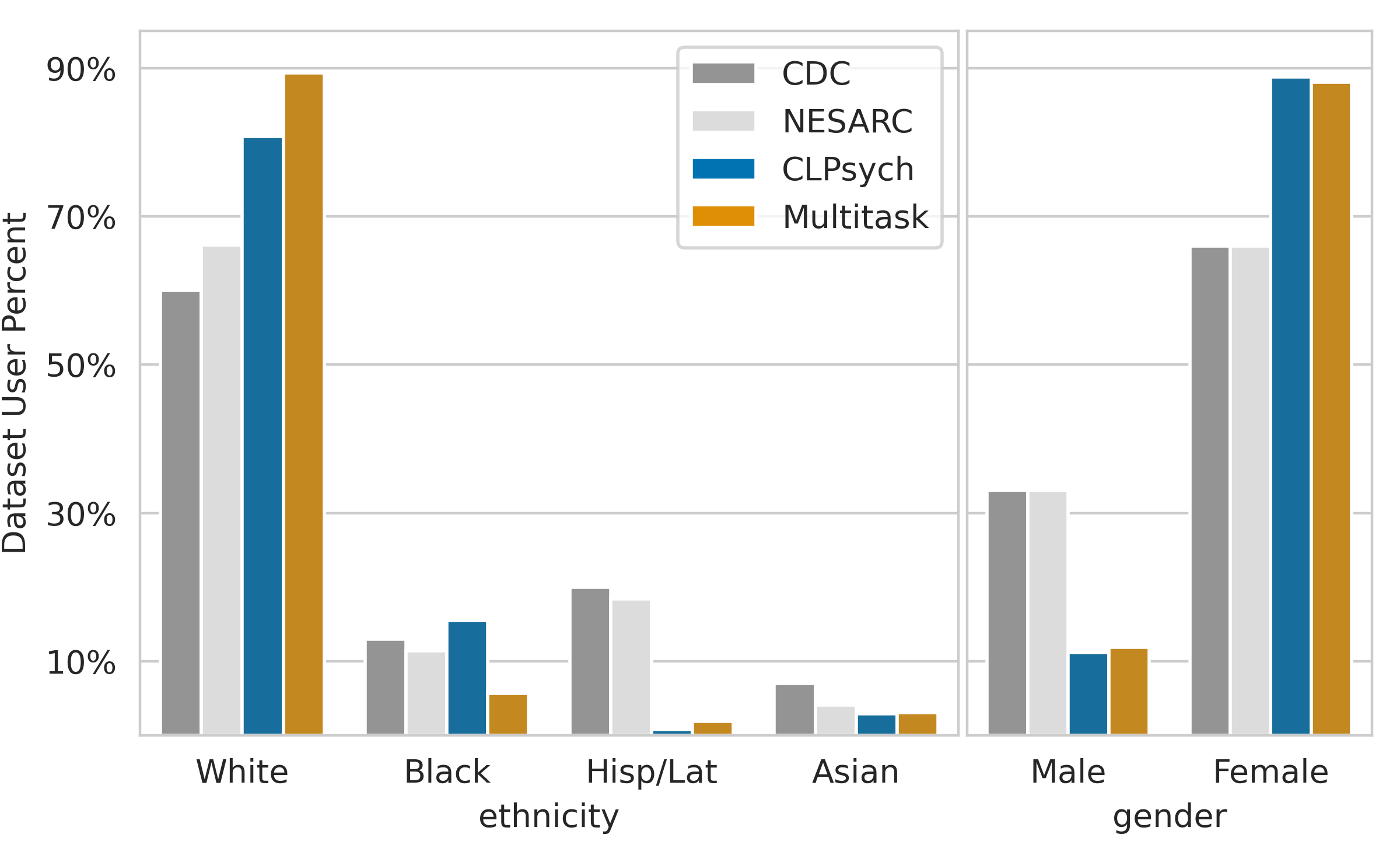}
  \caption{Hispanic/Latinx and Male individuals are underrepresented in both of our Twitter datasets (\proc{CLPsych} \& \proc{Multitask}) compared to the Twitter US population estimates (\proc{CDC} \& \proc{NESARC}).}
  \label{fig:DepressionPopulation}
\end{figure}

We combine the Twitter population estimates with depression rates in the US to get the target distributions of demographic users that we expect to observe in our datasets. \cref{fig:DepressionPopulation} shows differences between the expected, representative distribution and our complete Twitter datasets.

\label{sec:ClpsychAndMultitaskPopulation}

{\bf Results.} Based on these estimates, are the depression datasets demographically representative?
\cref{fig:DepressionPopulation} shows that \proc{CLPsych} and \proc{Multitask} are {\bf not demographically representative} with respect to either gender or race/ethnicity.
White individuals are over-represented, while Hispanic/Latinx individuals are the most underrepresented. In fact, there are \textit{no} male H/L individuals represented in the train split of \proc{CLPsych}.
\proc{Multitask} exhibits a larger population bias against minorities compared to \proc{CLPsych}; White individuals are over-represented and Black individuals are under-represented. 
With respect to gender, both \proc{CLPsych} and \proc{Multitask} have similar distributional skews -- females are over-represented compared to the depression adjusted general US population. At the user level, we found no major differences on number of tweets and vocabulary size between demographics: see details in \cref{apx:TweetAndVocabDemographics}.

Overall, \proc{CLPsych} and \proc{Multitask} are {\bf not demographically representative} with respect to US depression rates projected on Twitter demographic estimates.

\subsection{Do depression classifiers perform similarly across demographic groups?}
\label{sec:PerformanceCLPsych}

We consider this question through experimentation on our datasets, \proc{CLPsych} and \proc{Multitask}.

\textbf{Methods.}
We train a depression classifier on \proc{CLPsych} and \proc{Multitask} datasets.

We follow standard pre-processing procedures and filter numeric values, username mentions, retweets and urls from the raw tweet text.
We use $\ell_2$-regularized logistic regression models for all of our experiments.
TF-IDF vectors are used to represent text in and across tweets, along with mean-pooled 200 dimensional GloVe embeddings pretrained with 2B tweets \citep{pennington2014glove}.
The vocabulary is pre-filtered per training, as each unigram must appear at least 10 times across all the individuals in traning data.

We also experimented with Linguistic Inquiry Word Count (LIWC) features, a closed-vocabulary English lexicon containing 64 categories (excluding punctuation categories), ranging from linguistic dimensions to psychological processes covering emotions and personal concerns, traditionally used in psychological studies \citep{pennebaker2007operator}.
In social media analysis, LIWC has been shown to contain signals for mental health disorders \citep{ireland-iserman-2018-within, wolohan-etal-2018-detecting, mitchell-etal-2015-quantifying}, including \proc{CLPsych} and \proc{Multitask}.

We also use features based on topic distributions learned via Latent Dirichlet Allocation (LDA) \citep{blei2003latent}, following its implementation for Twitter data as specified in \citet{mitchell-etal-2015-quantifying} where all the tweets for an individual are combined into a ``document'' and we infer ``topics'' ($K=50$ topics).
All of the models in our experiments use all four feature groups: TF-IDF, GloVe embeddings, LIWC and LDA.
We considered using demographic labels as features, which have been shown to capture signals for depression in Twitter \citep{pietro-etal-2015-role}, but found no significant impact on our analysis or model performance; since demographic labels are not normally available, we do not include them in our analysis. See \cref{apx:Models} for further implementation details.

To measure the performance bias across demographic groups we report performance on each demographic group. 
However, the racial/ethnic minority groups in the data are vastly underrepresented.
While we address this by combining them into a `persons of color' (PoC) category, the PoC group is still small and limits the reliability and extension of our analysis in this data.

For each dataset, we randomly sample individuals with repetition to construct a training set (bootstrap method) and subsequently obtain a distribution of F1 scores (100 repetitions) followed by one way ANOVA and pairwise T-Tests for each demographic group pair.
Motivated by Simpson's Paradox \citep{blyth1972simpson} and the Matrix of Domination \citep{costanza2018design}, we combine the gender and race/ethnicity labels to create a matrix of demographics and report mean F1 scores and 95\% confidence interval of each demographic subcategory.
Additionally, we seek a metric to measure fairness in performance across demographic groups --- our criterion is that model performance should be independent of the demographic labels.
\citet{hardt2016equality} introduce \textit{equal odds} and \textit{equal opportunity}, two criteria that seek to equalize the FPR and TPR, or just FPR for the latter, across the protected attributes --- these are also known as `error rate balance' \cite{chouldechova2017fair}, `conditional procedure accuracy equality` \cite{berk2018fairness} and `classification parity' \cite{corbett2018measure}.
We compute the average pairwise equal odds and equal opportunity difference, a score of 0 means overall fairness, across the demographic groups in our boostrap sampling splits and report 95\% confidence interval.

\begin{table}[t]
\resizebox{\columnwidth}{!}{
\begin{tabular}{rrrlrlr}
  &  & \multicolumn{2}{c}{\proc{CLPsych}} & \multicolumn{2}{c}{\proc{Multitask}} \\
\cmidrule(lr){3-4} \cmidrule(lr){5-6} \\
\multirow{2}{*}{\bf Female}    & \bf White & \it 0.77 & \it $\pm$ 0.005 &  0.84 & $\pm$ 0.002\\
                          & \bf PoC   &  0.41 & $\pm$ 0.013 &  0.91 & $\pm$ 0.003 \\
\cmidrule(lr){3-6}
\multirow{2}{*}{\bf Male} & \bf White & \it 0.74 &  \it $\pm$ 0.008 &  0.83 & $\pm$ 0.005 \\
                          & \bf PoC  & \it 0.76 & \it $\pm$ 0.035 &  0.45 & $\pm$ 0.016 \\
\cmidrule[\heavyrulewidth](lr){1-6}
\multicolumn{2}{r}{\bf Equal Odds} & 0.21 & $\pm$ 0.023 & 0.13 & $\pm$ 0.013 \\
\multicolumn{2}{r}{\bf Opportunity} & 0.25 & $\pm$ 0.039 & 0.18 & $\pm$ 0.010\\
\end{tabular}}
\caption{Avg. F1 with 95\% conf. interval from bootstrap across gender and ethnicity groups {\it (italics: not significant)}, and avg. equal odds and equal opportunity differences. Models underperform for PoC in general, sometimes male PoC (\proc{Multitask}) or female PoC (\proc{CLPsych}).}
\label{tab:clpsych-merged-performance}
\end{table}

\textbf{Results.}
\cref{tab:clpsych-merged-performance} shows performance of classifiers trained on \proc{CLPsych} and \proc{Multitask} by demographic group.
Models trained on \proc{CLPsych} tend to perform worse on female PoC users compared to all other demographic groups.
While we observe higher model performance for \proc{Multitask} in general, models trained on \proc{Multitask} tend to perform worse on male PoC users, compared to all other demographic groups.
\proc{CLPsych} is scored worse with the fairness metrics compared to \proc{Multitask}.

In short, we observe that depression classifiers perform {\bf worse on people of color}, specifically female PoC in \proc{CLPsych} and male PoC \proc{Multitask}.

\subsection{Can  we  mitigate  demographic  biases  by changing characteristics of the dataset?}

Why do depression classifiers perform worse/inconsistently for PoC individuals? We conduct two analyses that investigate how the datasets may cause disparities in fairness.

\subsubsection{Data Size}
\label{sec:sizePerformance}
Perhaps the classifier performs worse on demographic groups because we have insufficient training data.
In \cref{sec:PerformanceCLPsych}, we observed fairer results with more data on \proc{Multitask} compared to \proc{CLPsych}.
We perform a dataset size experiment to verify the effect on model performance across demographics.

\textbf{Methods.}
How do results change with increased amounts of training data? To evaluate this gradient, we consider sampling an equal number of individuals from each demographic group and gradually increase overall dataset size until all available individuals available have been considered. At each dataset size step, we employ a similar bootstrap procedure to the one discussed in \cref{sec:PerformanceCLPsych}, sampling from the available user pool and training a classifier 25 times before moving on to the next dataset size. We continue adding data after a demographic group has been fully saturated to understand how information from overly-represented groups can generalize to under-represented groups.

\begin{figure}
  \includegraphics[width=\columnwidth]{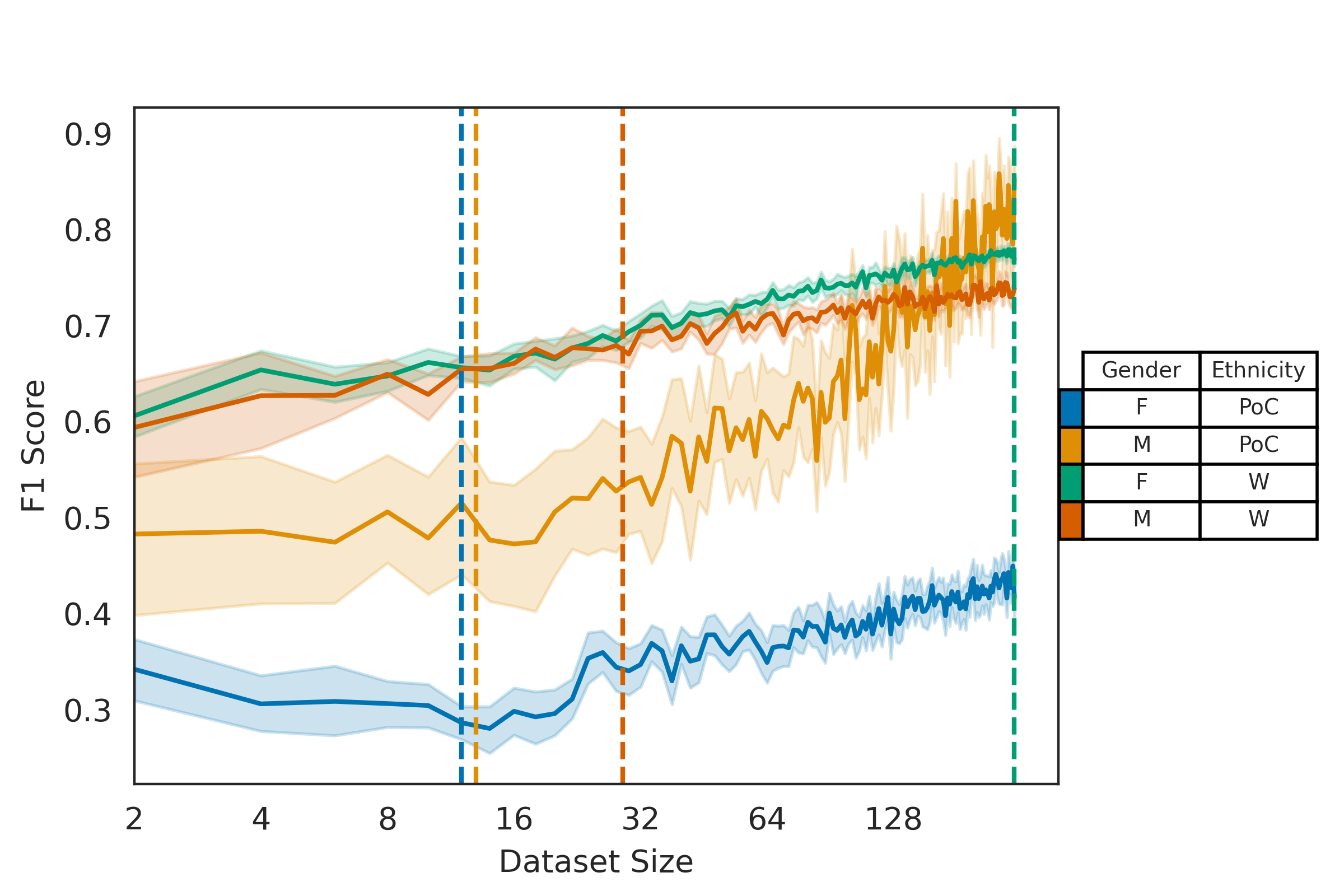}
  \caption{Log learning curve of varying dataset sample size on \proc{CLPsych}. Dashed vertical lines represent the total size of the demographic group. Curves suggest more data is helpful to close model performance gap between demographic groups.}
  \label{fig:LearningCurve}
\end{figure}

\textbf{Results.}
\cref{fig:LearningCurve} shows models performance as we increase training dataset size within the \proc{CLPsych} dataset; results for \proc{Multitask} are included in \cref{apx:MultitaskSize} and lead us to similar conclusions. As expected, overall performance across all classes improves with additional training data. Interestingly, even when the same amount of data is present for each demographic group, error rates remain higher for PoCs than for White users. This suggests that other factors beyond superficial representation are to blame for model degradation. It is also worthwhile noting that performance continues to improve for underrepresented groups after they have been fully saturated, thus implying that at least some signal generalizes between demographics.

\subsubsection{Data Balance}
\label{sec:balancePerformance}
What other factors could account for the difference in model performance on PoC? Below, we examine the effect of balancing the training data for the demographic groups.

\textbf{Methods.}
We consider \proc{Multitask} when constructing demographically-balanced datasets.
As explored in \cref{sec:DemographicBalance}, there are two different estimates of depression rates in demographic groups: \proc{CDC} and \proc{NESARC}.
We balance \proc{Multitask} to match the depression rates of both estimates, and name the models trained on those datasets \proc{Multitask CDC} and \proc{Multitask NESARC} respectively.
Additionally, we compare these with an \textit{even} balanced distribution.
Models are trained following the methodology in \cref{sec:PerformanceCLPsych}.

\begin{table}[t]
\resizebox{\columnwidth}{!}{
\begin{tabular}{rrrlrlrlrl}
  &  & \multicolumn{8}{c}{\proc{Multitask}} \\
\cmidrule(lr){3-10}
  &  & \multicolumn{2}{c}{\bf full} & \multicolumn{2}{c}{\bf \proc{NESARC}} & \multicolumn{2}{c}{\bf \proc{CDC}} & \multicolumn{2}{c}{\bf even} \\
\cmidrule(lr){3-4} \cmidrule(lr){5-6} \cmidrule(lr){7-8} \cmidrule(lr){9-10}
\multirow{2}{*}{\bf Female}    & \bf White & 0.84 & $\pm$ 0.002  & 0.75 & $\pm$ 0.006 &  0.75 & $\pm$ 0.008  &  0.69 & $\pm$ 0.013 \\
                               & \bf PoC   & 0.91 & $\pm$ 0.003 &  0.87 & $\pm$ 0.005 &  0.86 & $\pm$ 0.005  &  0.82 & $\pm$ 0.009 \\
                          \cmidrule(lr){3-10}
\multirow{2}{*}{\bf Male}      & \bf White & 0.83 & $\pm$ 0.005 &  0.74 & $\pm$ 0.007 &  0.74 & $\pm$ 0.008 &  0.68 & $\pm$ 0.011\\
                              & \bf PoC    & 0.45 & $\pm$ 0.016 &  0.56 & $\pm$ 0.029 &  0.54 & $\pm$ 0.033 &  0.48 & $\pm$ 0.030\\
\cmidrule[\heavyrulewidth](lr){1-10}
\multicolumn{2}{r}{\bf Equal Odds}   & 0.13 & $\pm$ 0.013 & 0.14  & $\pm$ 0.021  & 0.14 & $\pm$ 0.019  &  0.12 & $\pm$ 0.014\\
\multicolumn{2}{r}{\bf Opportunity}   & 0.18 & $\pm$ 0.010 & 0.16  & $\pm$ 0.032  & 0.18 & $\pm$ 0.036  &  0.12 & $\pm$ 0.027\\
\end{tabular}}
\caption{Avg. F1 with 95\% conf. interval from bootstrap across gender and ethnicity groups, and absolute avg. equal odds and equal opportunity differences. Balanced models close the performance difference gap at the cost of overall model performance.}
\label{tab:balance-performance}
\end{table}

\textbf{Results.}
\cref{tab:balance-performance} shows the average F1 score of  classifiers across gender and race/ethnicity groups.
We copy \proc{Multitask} column from \cref{tab:clpsych-merged-performance} (labeled as {\it full}) for ease of comparison.
There is a performance difference between male PoC users and the rest of the groups in models trained on \proc{Multitask} balanced datasets, similar to \proc{Multitask} full.
However, the performance difference is smaller on models trained on balanced datasets.
We observe no difference between balancing datasets according to \proc{NESARC} or \proc{CDC}, despite the 1.25x White user population increase in \proc{CDC}.
While fairness performance of both \proc{NESARC} or \proc{CDC} are similar to the full dataset, the \textit{even} dataset shows considerable improvement for both fairness metrics at the cost of model performance.

Our experiments with both dataset size and balance show that it matters when datasets are not demographically representative, and as shown in section \ref{sec:DemographicBalance}, they are not.

\subsection{Can differences in features between demographic groups account for classifier biases?}
\label{sec:featureDiscrepancies}
We have demonstrated a demographic bias in classifiers trained on \proc{CLPsych} and \proc{MultiTask}. Perhaps differences in feature representations between the groups can explain some of this bias. 
We examine LIWC features, which previous research has identified as useful in depression classification \citep{glen-leary, coppersmith-etal-2014-quantifying}, in addition to performance analysis in \cref{appx:FeatureStudy}.

{\bf Methods.} Previous research using the \proc{CLPsych} and \proc{Multitask} datasets has identified LIWC dimensions that over-index amongst depressed individuals: 
\textit{Negative Emotion} (negemo), \textit{Swearing} (swear), \textit{Anger} (anger), \textit{Anxiety} (anx) and \textit{First-person Pronoun Usage} (Pro1) \cite{coppersmith-etal-2014-quantifying}.\footnote{\textit{Pro1} is constructed by combining the \textit{i} and \textit{we} LIWC categories.} We evaluate whether this finding holds within each demographic group independently and whether there exist shifts between demographic groups.

\begin{figure}
  \includegraphics[width=\columnwidth]{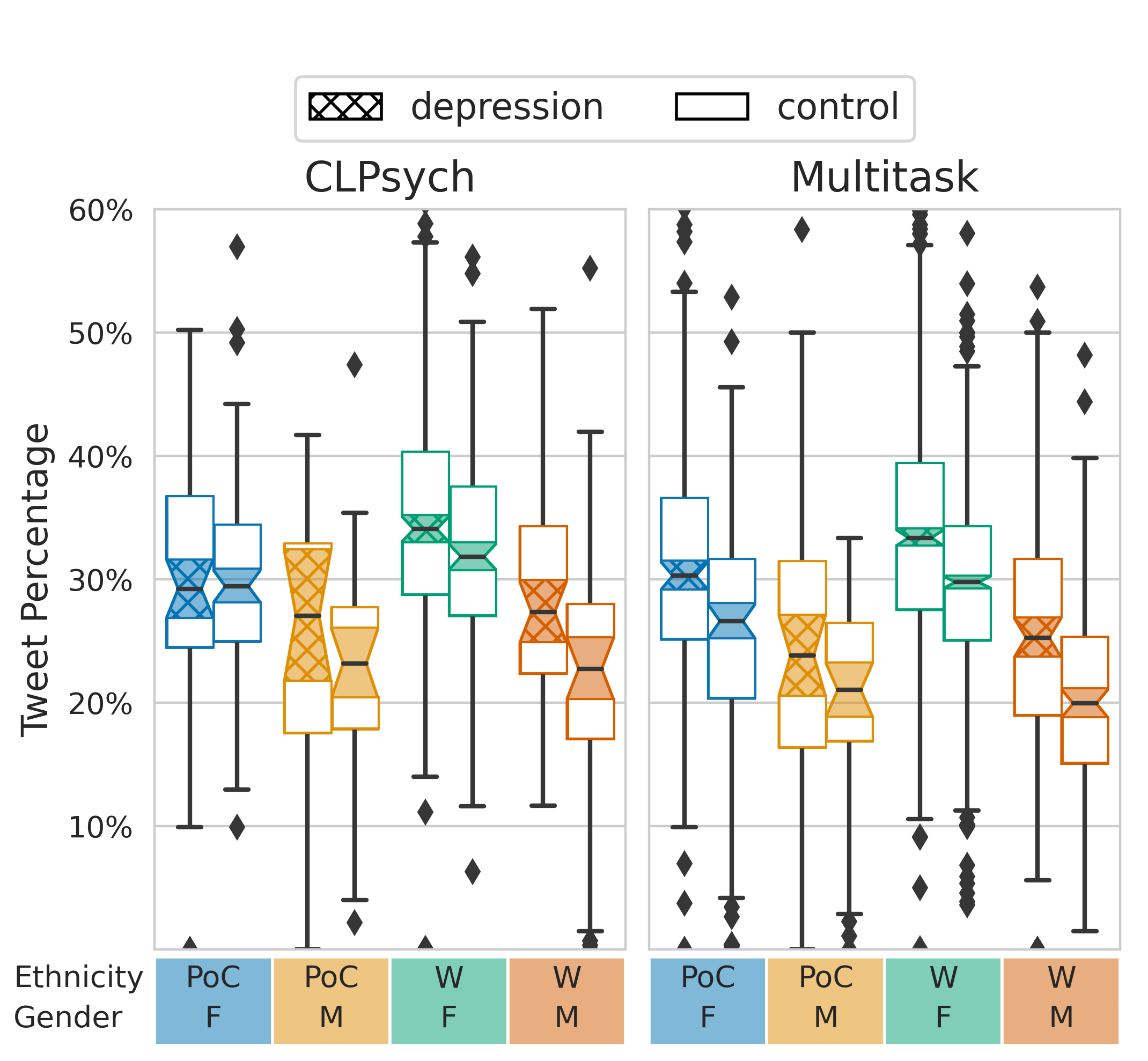}
  \caption{Percent of tweets containing first-person pronoun \textit{(Pro1)} previously shown to correlate with the depression group; shaded notches show median CI (95\%) with bootstrap $n=10000$. Correlations shown by other work are not universal among demographic groups.}
  \label{fig:liwc}
\end{figure}

{\bf Results.} \cref{fig:liwc} shows the distribution over users of percentage of tweets with at least one word matching the \textit{Pro1} category across demographic groups, with shaded notches showing median confidence interval. Results for other LIWC categories associated with depression are similar to those for \textit{Pro1} (see \cref{sec:appendix_figs}).

From previous research, we expect to observe a greater \textit{Pro1} prevalence in depression groups compared to controls across all demographics i.e. shaded notches of depression box should not overlap with control in each demographic category.
However, in \proc{CLPsych}, we do not observe any difference in prevalence of \textit{Pro1} in the PoC groups, and in \proc{Multitask} we do not observe any difference in prevalence in the male PoC group, both contradicting previous results.
We also observe a correlation between prevalence of these LIWC categories in the depression group and downstream model performance for each demographic group, corroborating previous findings of the correlation of LIWC categories and depression signals \citep{glen-leary, coppersmith-etal-2014-quantifying}.
In general, female groups for both the control and depression sets tend to have a higher prevalence of \textit{Pro1} compared to their male counterparts, suggesting a difference in language between the groups.

In short, LIWC correlations with depression are not universal across demographic groups. Furthermore, a closed vocabulary feature, such as LIWC, may contribute towards bias against some demographic groups.

\section{Limitations}
\label{sec:caveats}

\textbf{Depression and Control Groups.} 
The method used to curate the depression group in these datasets is susceptible to \textit{self-selection} bias, as noted by \citet{coppersmith-etal-2014-quantifying} and \citet{amir-etal-2019-mental}, as it likely over-represents individuals who are more vocal about their condition. Therefore, differences in use of social media and cultural perceptions around mental health may introduce biases in these datasets. Further, while expert annotators identified non-genuine disclosures depression and removed these individuals from the \proc{CLPsych} and \proc{Multitask} datasets, they did not verify the authenticity of the diagnosis. Similarly, individuals in the control group may have been actually diagnosed with depression, but did not disclose their condition anywhere in their public timeline. Thus, labels for both the the depression and control groups are bound to be noisy. 

\textbf{Representation.} 
We balance the \proc{Multitask} dataset to match depression rates in the US, which may not be representative of non-US populations.  
Additionally, we preserve the even depression/control splits for class balance in model training, instead of using the true depression/control population rates (about 1/10).
These splits are not reflective of the true depression prevalence and their use may need to be modified depending on downstream classifier use.

\textbf{Demographic Labels.}
Due to dataset limitations, the ethnicity and gender labels for this study were inferred using the unigram model from \citep{wooddoughty2020using}. 
This model considers only the four largest race/ethnicity groups in the US, which aligns with conventions from \citet{CDC} but ignores smaller populations and multi-racial categories. Further, in some of our analyses, we combine Asian, Black and Hispanic/Latinx individuals into people of color due to a lack of data.
With respect to gender, the male and female labels used by this model do not consider individuals who fall outside of traditional binary gender. 
As our experiments rely on upstream demographic label inference, we cannot fully rule out confounding factors due to e.g. noisy labels in our experimentation, but we perform a high-confidence filter on demographic labels and statistical testing on results to strengthen our conclusions.

\section{Conclusion and Recommendations}
We examine whether datasets and the resulting trained classifiers for depression prediction are fair across demographic groups.
Our analysis finds that (1) depression datasets are {\bf not} demographically representative, in some cases excluding entire intersectional groups and (2) the resulting classifiers perform worse on people of color in general. 
In examining the reason for these differences, we find that performance difference could be improved after accounting for (3) the size of the dataset and balance across demographic groups. (4) Finally, we show that signals of depression found by previous work using e.g. LIWC features are \textit{not} equally representative for all demographics.

These findings should give pause to researchers in this area. Since datasets and the resulting models are not demographically representative, advances in methods may be furthering biases towards some groups. Worse, since some intersectional demographic groups are not even represented in the data, compounded by the fact that most datasets do not have labels for demographic groups, we currently lack the means to even check how new methods perform on each group. Going forward, research in this area should include demographic analyses so that improvements on the overall dataset can be contextualized by how they perform on each demographic group.

At the same time, there is reason for optimism. Our data balancing and dataset size experiments reduced demographic disparities of trained models. This suggests that research can continue with existing datasets, but with the modifications we proposed.
We release the demographically balanced dataset from our experimentation upon appropriate terms of usage agreement.

Ultimately, the best approach will be to construct new datasets that better represent the population, especially underrepresented minorities who are most at risk from systematic bias.
This may necessitate changes to the data collection methods themselves, which may bias collection against certain groups. For example, self-reports may be problematic as they rely on cultural attitudes towards the expression of mental health information. Further research is needed to understand if self-reports and other proxy-based methods for obtaining labels can be successfully adapted to include a more diverse population, e.g. do keywords used to collect tweets skew resulting user populations? Further, to produce more conclusive insights with respect to demographics, language-based classifiers for demographic labels need to be further improved. Alternatively, other data collection strategies, such as the cohort method of \citet{amir-etal-2019-mental}, may be more successful at ensuring representative datasets.

\section*{Acknowledgments}
The authors gratefully acknowledge Elizabeth Salesky, Zach Wood-Doughty, Rachel Wicks, Alexandra DeLucia and the CLSP NLP reading group for helpful feedback, and we thank the anonymous reviewers for their helpful comments.

\bibliographystyle{acl_natbib}
\bibliography{anthology,eacl2021}

\clearpage
\newpage
\appendix
\onecolumn
\label{sec:appendix}
\renewcommand{\thepage}{} 

\section{Demographic Labels Analysis} \label{apx:HighConfidenceFilter}
Due to data anonymization, we must utilize content-based demographic classifiers to infer labels. This introduces noise due to classification error to our analyses. In order to reduce these effects we consider 3 techniques: high confidence filter, name labels, and tweet sampling.

\textbf{High Confidence Filter.} We select users whose most probably demographic class for both gender and race is $> 0.95$ probability.

\textbf{Random Sampling.} Considers all available users; randomly split each individual’s tweets into  two  independent  pools  so  that  demographic  and mental health inferences are based on separate sets of data.

\textbf{Name Labels.} With the permission of \citet{coppersmith-etal-2014-quantifying}, we were able to collect name attributes for 622 of individuals in \proc{CLPsych}. To obtain demographic labels from name attributes we used the demographer's neural name classifier \citep{wood-doughty-etal-2018-predicting}, and ethnic/race name classifier \citep{wooddoughty2020using}. Due to the small number of individuals that we could obtain name attributes, our analysis of this technique is limited.

\begin{figure}[hb]
  \includegraphics[width=\columnwidth]{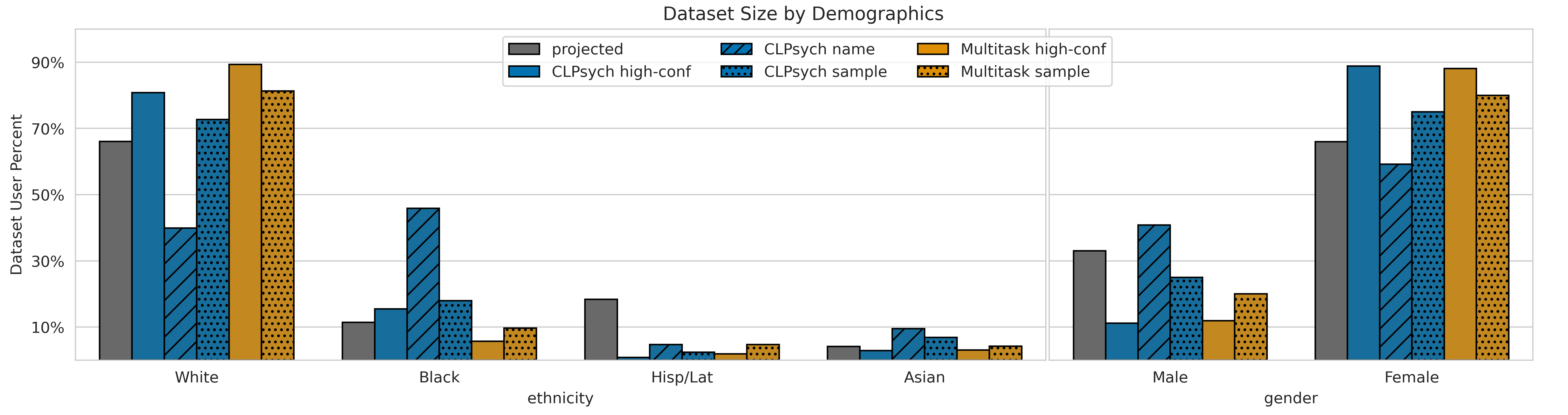}
  \caption{Hispanic/Latinx and Male individuals are underrepresented in both of our Twitter datasets (\proc{CLPsych} \& \proc{Multitask}) compared to the Twitter US population estimates (\proc{NESARC}) in all our labeling techniques.}
  \label{fig:DepressionPopulationExtended}
\end{figure}

\cref{fig:DepressionPopulationExtended} shows the population percentage of the three techniques compared to the projected distribution. For all the techniques, some trends still hold, mainly Hispanic/Latinx group is underrepresented.
Additionally, for the high confidence and split techniques, White and female groups are overrepresented and male underrepresented. 
In contrast, in the name based approach, the Black and male groups are overrepresented while the White and female groups are underrepresented.

While the name based technique shows more promising data distributions, the Hispanic/Latinx group is still vastly underrepresented. However, the rest of the demographic groups seem more fair and closer (sometimes better) than the target distribution, but do these distributions translate to greater downstream performance?

\cref{tab:high_confidence_table} shows the performance of mental health models trained on the different datasets obtained from our demographic labeling techniques. High confidence filtering has the largest performance difference and higher fairness metrics compared to random sampling and name based approaches. However, we still observe similar trends with PoC groups performing in general lower than White groups.

\begin{table}[hb]
\centering
\resizebox{.7\columnwidth}{!}{
\begin{tabular}{rrrlrlrlrlrl}
  &  & \multicolumn{6}{c}{\proc{CLPsych}} & \multicolumn{4}{c}{\proc{Multitask}} \\
\cmidrule(lr){3-8} \cmidrule(lr){9-12}
  &  & \multicolumn{2}{c}{\bf High Conf.} & \multicolumn{2}{c}{\bf Rand. Sample} & \multicolumn{2}{c}{\bf Name Labels} & \multicolumn{2}{c}{\bf High Conf.} & \multicolumn{2}{c}{\bf Rand. Sample} \\
\cmidrule(lr){3-4} \cmidrule(lr){5-6} \cmidrule(lr){7-8} \cmidrule(lr){9-10} \cmidrule(lr){11-12}\\
\multirow{2}{*}{\bf Female}    & \bf White & \it 0.77 & \it $\pm$ 0.005 &  0.77 & $\pm$ 0.005 &  0.72 & $\pm$ 0.006  &  0.84 & $\pm$ 0.002 &  0.84 & $\pm$ 0.003 \\
                               & \bf PoC   & 0.41 & $\pm$ 0.013         &  0.47 & $\pm$ 0.020 &  0.44 & $\pm$ 0.018  &  0.91 & $\pm$ 0.003 &  0.87 & $\pm$ 0.010 \\
                          \cmidrule(lr){3-12}
\multirow{2}{*}{\bf Male}      & \bf White & \it 0.74 & \it $\pm$ 0.008 &  0.74 & $\pm$ 0.014 &  0.48 & $\pm$ 0.032 &  0.83 & $\pm$ 0.005  &  0.78 & $\pm$ 0.011 \\
                              & \bf PoC    & \it 0.76 & \it $\pm$ 0.035 &  0.72 & $\pm$ 0.040 &  0.56 & $\pm$ 0.042 &  0.45 & $\pm$ 0.016  &  0.78 & $\pm$ 0.033 \\
\cmidrule[\heavyrulewidth](lr){1-12}
\multicolumn{2}{r}{\bf Equal Odds}         & 0.21 & $\pm$ 0.023         & 0.13  & $\pm$ 0.014  & 0.19 & $\pm$ 0.023  &  0.13 & $\pm$ 0.013 & 0.13 & $\pm$ 0.021\\
\multicolumn{2}{r}{\bf Opportunity}        & 0.25 & $\pm$ 0.039         & 0.16  & $\pm$ 0.022  & 0.18 & $\pm$ 0.031  &  0.18 & $\pm$ 0.010 & 0.14 & $\pm$ 0.034\\
\end{tabular}}
\caption{Avg. F1 with 95\% conf. interval from bootstrap across gender and ethnicity groups, and avg. equal odds and equal opportunity differences {\it (italics: not significant)}. Models underperform for PoC in general, sometimes male PoC (\proc{Multitask}) or female PoC (\proc{CLPsych}).}
\label{tab:high_confidence_table}
\end{table}

\clearpage
\newpage

\section{Tweets and Vocabulary in Demographics} \label{apx:TweetAndVocabDemographics}
\begin{figure}[h]
\begin{minipage}{\linewidth}
\centering
  \includegraphics[width=\linewidth]{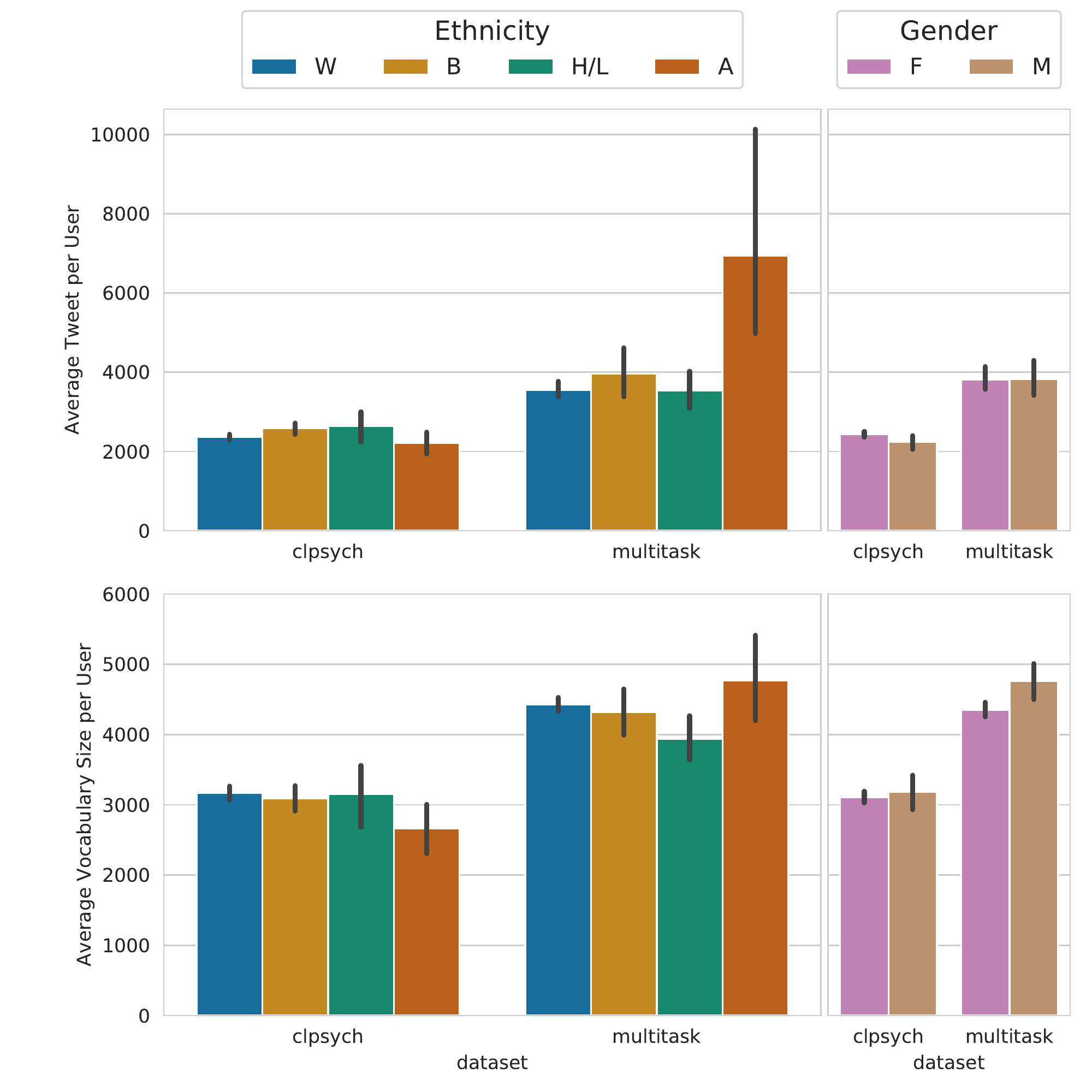}
  \caption{Average tweet number per user per demographic category, as well as average vocabulary size. Black bars show 95\% CI.}
  \label{fig:tweet_and_vocab_demographics}
\end{minipage}
\hfill
\end{figure}

\cref{fig:tweet_and_vocab_demographics} shows the distribution of the average number of tweets and vocabulary size across all our demographic groups. There is no statistical significant difference in the means between demographic groups within the datasets. Very high variance is observed in the Asian group in multitask, although their vocabulary size is not as high variance. Additionally, \proc{Multitask} dataset has on average higher number of tweets per user (as they were not limited to 3000 as in \proc{CLPsych}) and consequently a higher average vocabulary size.

\clearpage
\newpage

\section{Model Specifications} \label{apx:Models}
\textbf{Tokenization.}
Raw text within in Tweets was tokenized using a modified version of the Twokenizer \cite{o2010tweetmotif}. English contractions were expanded, while specific retweet tokens, username mentions, URLs, and numeric values were replaced by generic tokens. As pronoun usage tends to differ in individuals living with depression \cite{vedula2017emotional}, we removed any English pronouns from our stop word set (English Stop Words from nltk.org). Case was standardized across all tokens, with a single flag included if an entire post was made in uppercase letters.

\textbf{Features}. Text from all documents for an individual are concatenated together and tokenized as previously described. The vocabulary of each training procedure is fixed to a maximum of 100-thousand unigrams selected based on KL-divergence of the class-unigram distribution with the class-distribution of stop words \cite{chang2012phillies}. This reduced bag-of-words representation is then used to generate the following additional feature dimensions: a 50-dimensional LDA topic distribution \cite{blei2003latent}, a 64-dimensional LIWC category distribution \cite{pennebaker2007operator}, and a 200-dimensional mean-pooled vector of GloVe embeddings \cite{pennington2014glove}. The reduced bag-of-words representation is transformed using TF-IDF weighting \cite{ramos2003using}.\footnote{All data-specific feature transformations (e.g. LDA, TF-IDF) are learned without access to development or test data.}

\textbf{Hyperparameter Selection.} 
Each model is trained using a hyperparameter grid search over the regularization strength \{1e-3, 1e-2, 1-e1, 1, 10, 100, 1e3, 1e4, 1e5\}, class weighting \{None, Balanced\}, and feature set standardization \{On, Off\}. Hyperparameters were selected to maximize held-out F1 score within a 20\%-sied held-out split of the training data.

\clearpage
\newpage

\section{Multitask Size Experiment}
\label{apx:MultitaskSize}

\begin{figure}[h]
\begin{minipage}{\linewidth}
\centering
  \includegraphics[width=\linewidth]{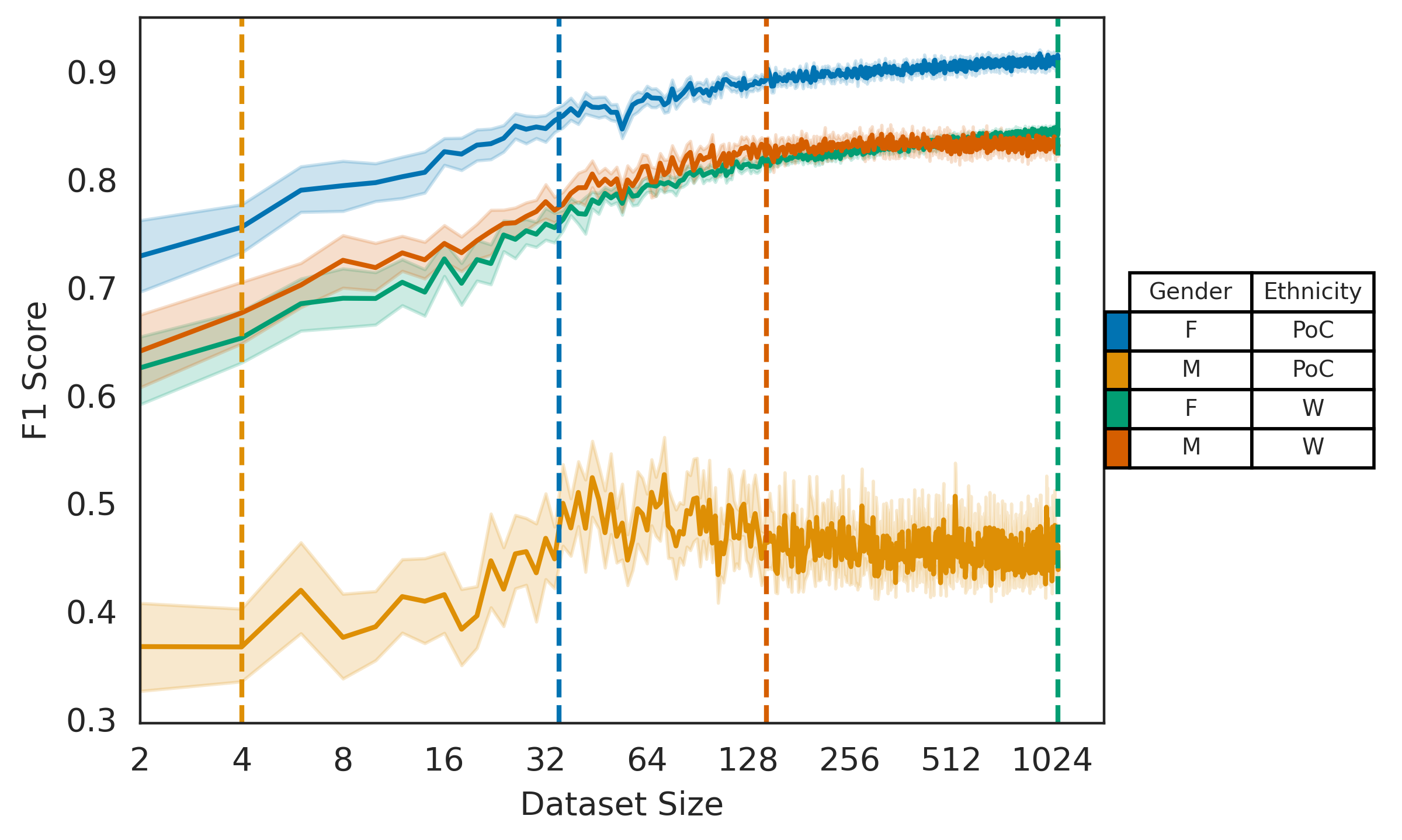}
  \caption{Log learning curve of varying dataset sample size on \proc{Multitask}. Dashed vertical lines represent the total size of the demographic group.}
  \label{fig:multitask_size}
\end{minipage}
\hfill
\end{figure}

\clearpage
\newpage

\section{Feature Study}
\label{appx:FeatureStudy}

We observed performance difference between demographic groups for both our datasets with mental health models using the following feature groups: LIWC, LDA, GloVe and TF-IDF as specified in \cref{apx:Models}. 
To explore the source of the performance difference observed in demographic groups, we train classifiers with each individual feature group and present average F1 score per demographic groups as well as our fairness metrics in \cref{tab:abliation}.

TF-IDF and GloVe embeddings yield better model performance than the other feature groups at the expense of fairness as measure by our fairness metrics. The most fair feature set was LIWC, although it was also the least informative feature set resulting in worst performing models.

However, trends observed in \cref{sec:PerformanceCLPsych} still apply in all feature groups, mainly models tend to underperform for PoC groups (female PoC groups in \proc{CLPsych}).

\begin{table}[h]
\resizebox{\columnwidth}{!}{
\begin{tabular}{rrrlrlrlrlrl}
  &  & \multicolumn{10}{c}{\proc{CLPsych}} \\
\cmidrule(lr){3-12}
  &  & \multicolumn{2}{c}{\bf full} & \multicolumn{2}{c}{\bf TF-IDF} & \multicolumn{2}{c}{\bf LDA} & \multicolumn{2}{c}{\bf LIWC} & \multicolumn{2}{c}{\bf GloVe} \\
\cmidrule(lr){3-4} \cmidrule(lr){5-6} \cmidrule(lr){7-8} \cmidrule(lr){9-10} \cmidrule(lr){11-12}
\multirow{2}{*}{\bf Female}    & \bf White & \it 0.77 & \it $\pm$ 0.005  & 0.77 & $\pm$ 0.005 &  0.70 & $\pm$ 0.008  &  0.68 & $\pm$ 0.010 &  0.71 & $\pm$ 0.007\\
                               & \bf PoC   & 0.41 & $\pm$ 0.013          & 0.42 & $\pm$ 0.001 &  0.41 & $\pm$ 0.026  &  0.45 & $\pm$ 0.020 &  0.47 & $\pm$ 0.020\\
                          \cmidrule(lr){3-12}
\multirow{2}{*}{\bf Male}      & \bf White & \it 0.74 & \it $\pm$ 0.008  & 0.74 & $\pm$ 0.008 &  0.65 & $\pm$ 0.019 &  0.68 & $\pm$ 0.021  &  0.68 & $\pm$ 0.019\\
                              & \bf PoC    & \it 0.76 & \it $\pm$ 0.035  & 0.74 & $\pm$ 0.034 &  0.62 & $\pm$ 0.054 &  0.55 & $\pm$ 0.040  &  0.75 & $\pm$ 0.032\\
\cmidrule[\heavyrulewidth](lr){1-12}
\multicolumn{2}{r}{\bf Equal Odds}    & 0.21 & $\pm$ 0.023 & 0.20  & $\pm$ 0.023  & 0.21 & $\pm$ 0.031  &  0.17 & $\pm$ 0.021 &  0.21 & $\pm$ 0.022\\
\multicolumn{2}{r}{\bf Opportunity}   & 0.25 & $\pm$ 0.039 & 0.25  & $\pm$ 0.040  & 0.24 & $\pm$ 0.039  &  0.17 & $\pm$ 0.028 &  0.18 & $\pm$ 0.029\\
\end{tabular}}
\caption{Avg. F1 with 95\% conf. interval from bootstrap across gender and ethnicity groups, and absolute avg. equal odds and equal opportunity differences. PoC groups (female PoC) perform worse for models separately trained on each of our feature group.}
\label{tab:abliation}
\end{table}

\clearpage
\newpage

\section{Additional LIWC Categories Figures}
\label{sec:appendix_figs}

Previous research has identified specific LIWC dimensions that are of importance in depression groups.
In addition to \textit{First-person pronoun} (pro1), categories like \textit{Negative Emotion} (negemo), \textit{Swearing} (swear), \textit{Anger} (anger) and \textit{Anxiety} (anx) have been shown to be more prominent in the depression groups compared to control.
We observe variation on prevalence on depression groups in all categories mentioned above among demographic groups, showing that LIWC features are \textit{not} equally representative for all demographics

\begin{figure}[h]
\begin{minipage}{.45\linewidth}
\centering
  \includegraphics[width=\linewidth]{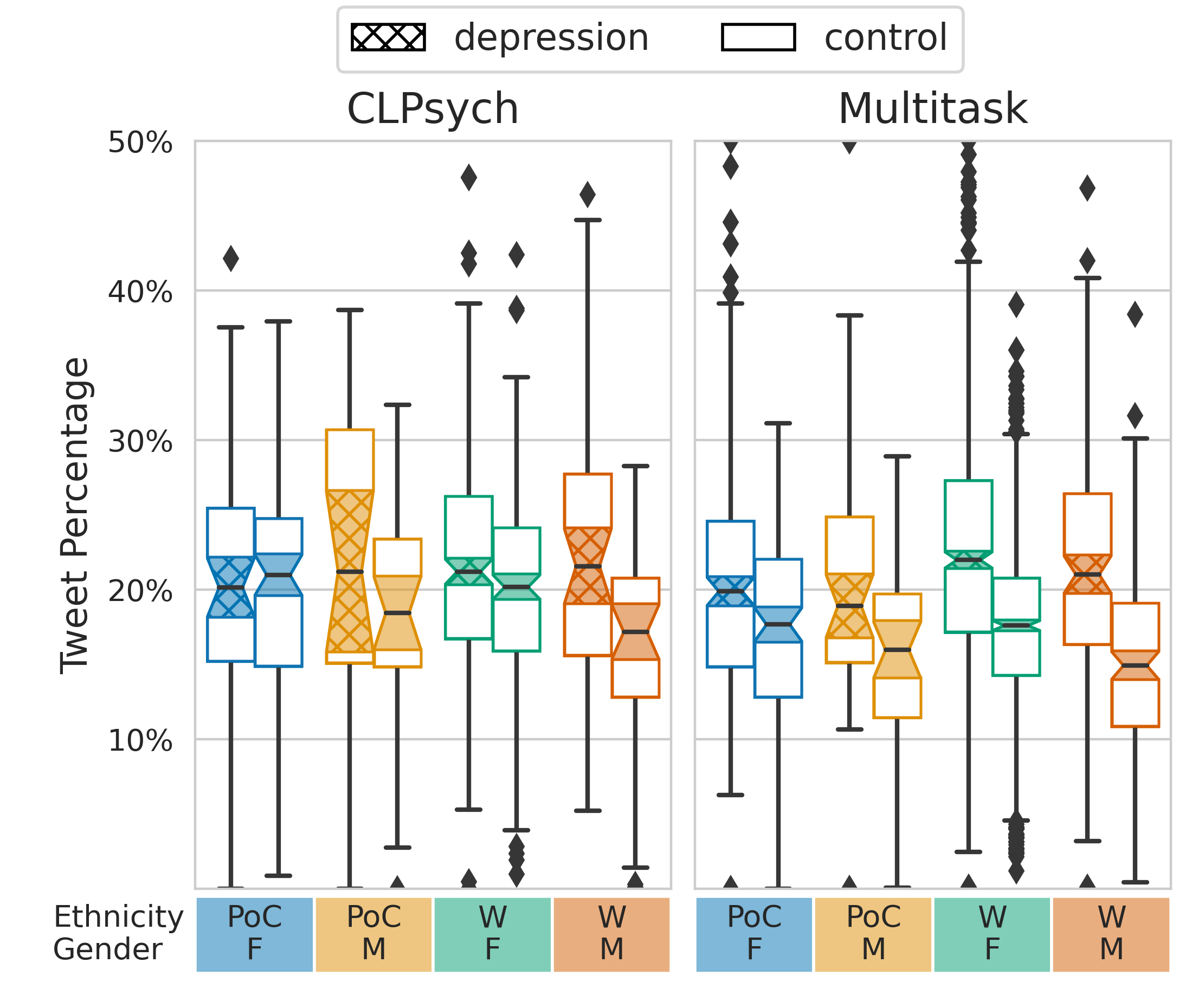}
  \caption{Negative Emotion Usage (\textit{negemo}) LIWC category representation within each individual, previously shown to correlate with the depression group; statistical significance marked by not overlapping shaded notches. We observe median statistical difference only for \textbf{White male} individuals in \proc{CLPsych} and \textbf{White groups} in \proc{Multitask}.}
  \label{fig:liwc-negemo}
\end{minipage}
\hfill
\begin{minipage}{.45\linewidth}
\centering
  \includegraphics[width=\linewidth]{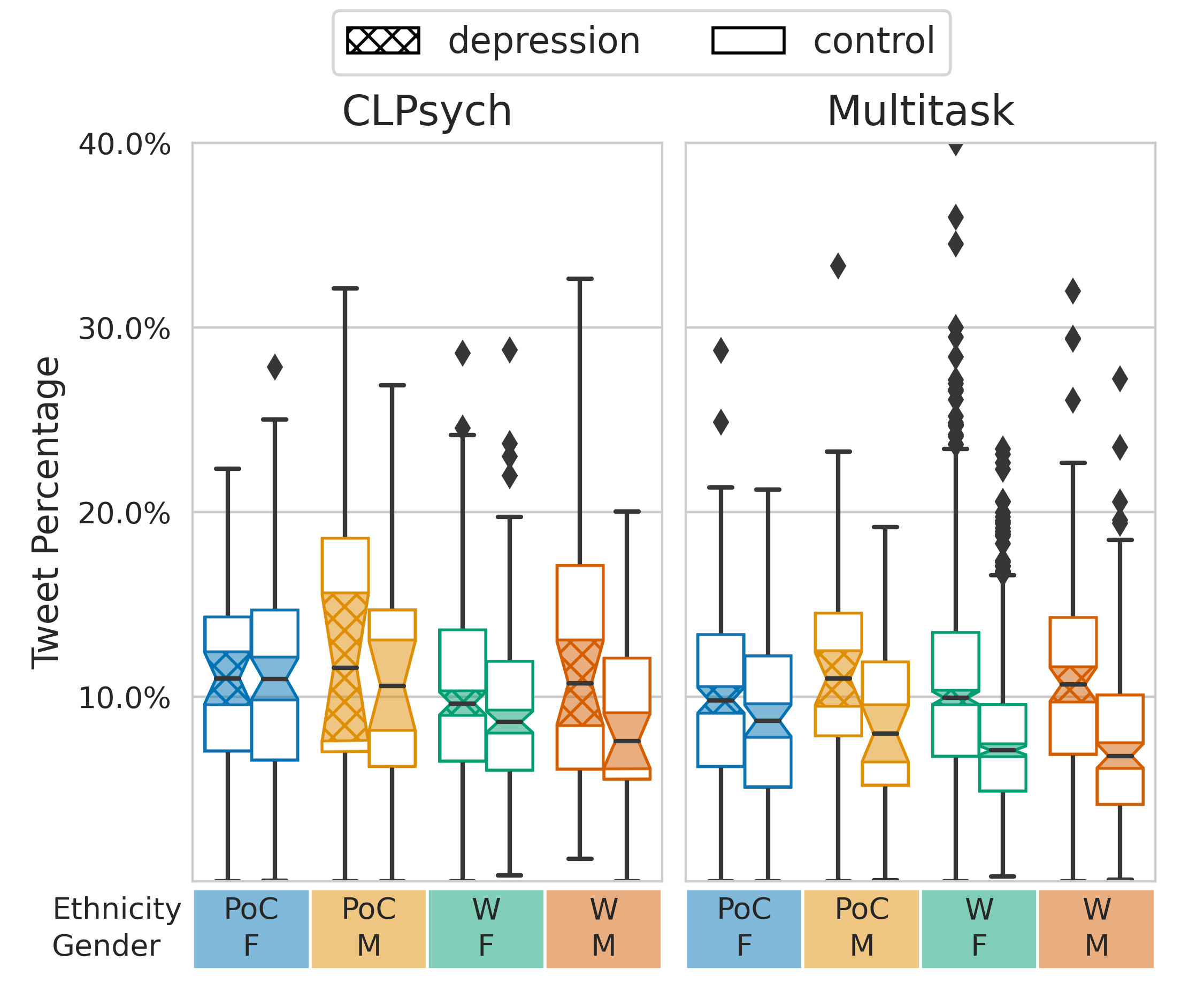}
  \caption{Anger (\textit{anger}) LIWC category representation within each individual, previously shown to correlate with the depression group; statistical significance marked by not overlapping shaded notches. We only observe median statistical difference for \textbf{no groups} \proc{CLPsych} and \textbf{White groups} in \proc{Multitask}}
  \label{fig:liwc-anger}
\end{minipage}
\end{figure}

\begin{figure}[h]
\begin{minipage}{.45\linewidth}
\centering
  \includegraphics[width=\linewidth]{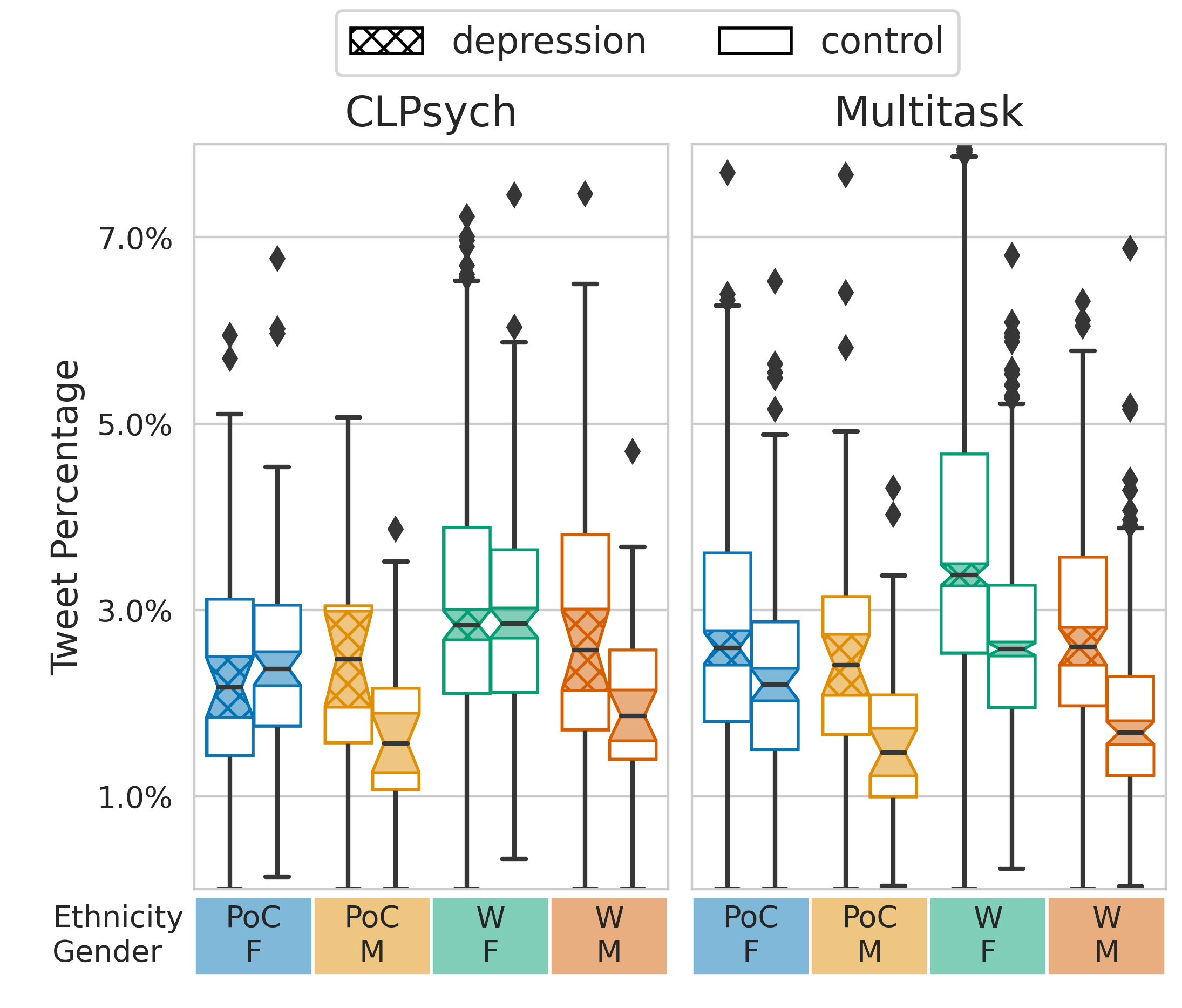}
  \caption{Anxiety (\textit{anx}) LIWC category representation within each individual, previously shown to correlate with the depression group; statistical significance marked by not overlapping shaded notches. We observe median statistical difference for the \textbf{male groups} in \proc{CLPsych} and \textbf{all} demographic categories in \proc{Multitask}}
  \label{fig:liwc-anx}
\end{minipage}
\hfill
\begin{minipage}{.45\linewidth}
\centering
  \includegraphics[width=\linewidth]{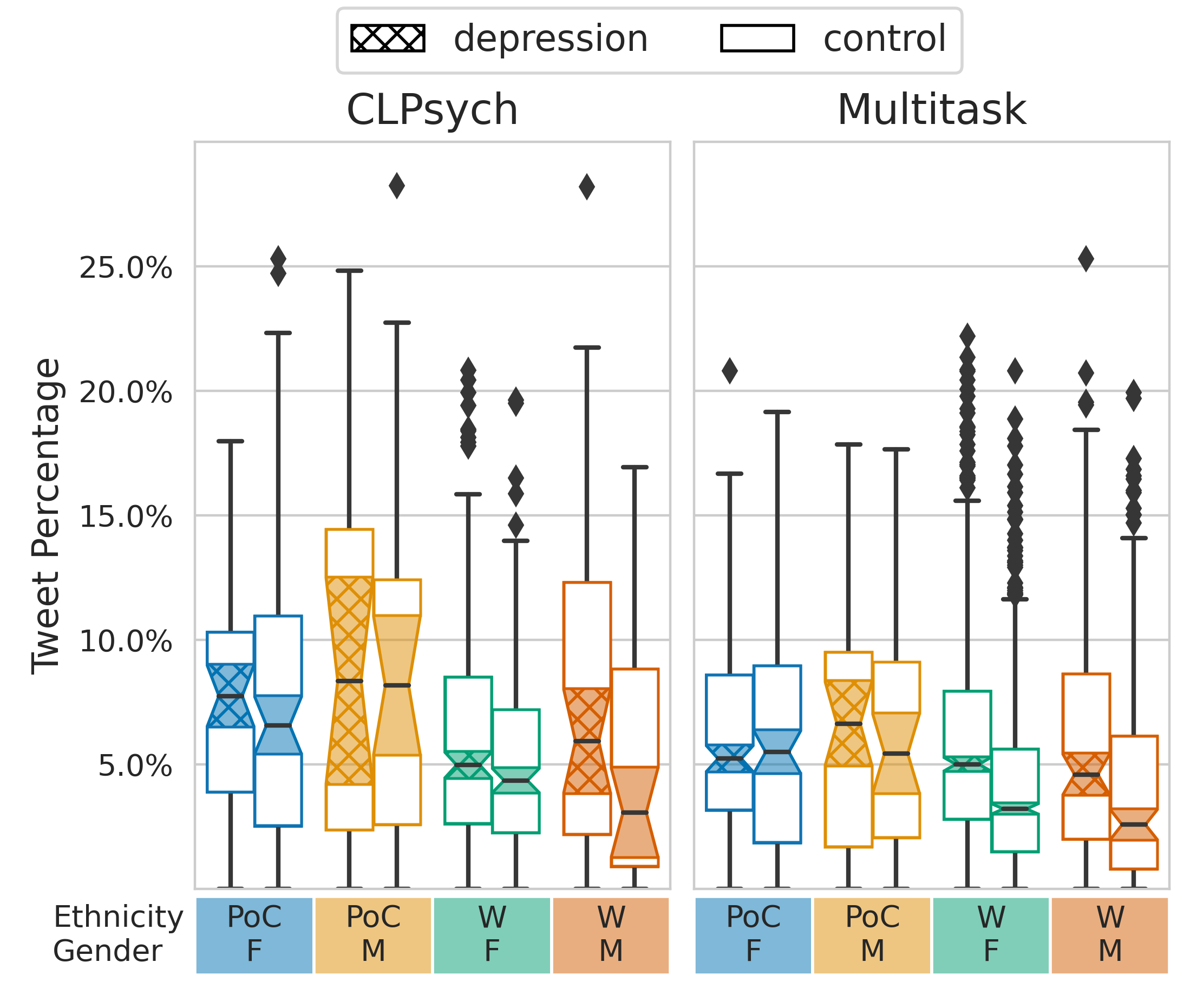}
  \caption{Swearing (\textit{swear}) LIWC category representation within each individual, previously shown to correlate with the depression group; statistical significance marked by not overlapping shaded notches. We observe median statistical difference for \textbf{no groups} in \proc{CLPsych} and \textbf{White groups} in \proc{Multitask}.}
  \label{fig:liwc-swear}
\end{minipage}
\end{figure}

\end{document}